\pdfoutput=1
\documentclass[10pt,twocolumn,letterpaper]{article}

\usepackage{cvpr}
\usepackage{times}
\usepackage{epsfig}
\usepackage[breaklinks=true,bookmarks=false]{hyperref}
\usepackage{capt-of}
\usepackage{color}
\usepackage{algpseudocode}
\usepackage{algorithm}

\usepackage{graphics, graphicx}
\usepackage{subfigure}
\usepackage{epstopdf}
\usepackage{float}
\usepackage{multirow}
\usepackage{cite}
\usepackage{latexsym,bm,amsmath,amssymb}
\usepackage{enumerate}
\usepackage[noblocks]{authblk}
\usepackage{flushend}

\cvprfinalcopy

\begin{document}

\title{Fusing Hierarchical Convolutional Features for Human Body Segmentation and Clothing Fashion Classification}

\author{Zheng Zhang, Chengfang Song, Qin Zou\thanks{Corresponding author.}  \\
School of Computer Science, Wuhan University, P.R.~China \\

E-mails: \{zhangzheng, songchf, qzou\}@whu.edu.cn
}

\maketitle

\begin{abstract}
The clothing fashion reflects the common aesthetics that people share with each other in dressing. To recognize the fashion time of a clothing is meaningful for both an individual and the industry. In this paper, under the assumption that the clothing fashion changes year by year, the fashion-time recognition problem is mapped into a clothing-fashion classification problem. Specifically, a novel deep neural network is proposed which achieves accurate human body segmentation by fusing multi-scale convolutional features in a fully convolutional network, and then feature learning and fashion classification are performed on the segmented parts avoiding the influence of image background. In the experiments, 9,339 fashion images from 8 continuous years are collected for performance evaluation. The results demonstrate the effectiveness of the proposed body segmentation and fashion classification methods.

\end{abstract}

\section{Introduction}
\label{sec:intro}
Fashion is a product of social psychology which reflects the common aesthetics that people share with each other. Clothing is probably one of the most salient and expressive medium to construct and deliver fashion concepts. In the fast-paced modem life, the clothing fashion updates year by year, and a clothing commonly attaches a fashion-year label which indicates the year it belongs to. The knowledge on clothing fashion can guide us to dress fashionable clothes and help us avoid buying outdated ones. However, dazzled by the abundant clothing in the shop, it is often difficult for human to remember and recognize the clothing's fashion features, not to mention identifying the fashion-year label. Therefore, it makes great sense to study automatic method to recognize the clothing fashion.

Using computer vision and machine learning techniques, we develop a clothing fashion recognition method and recognize the fashion-year label of a clothing automatically. This method is built upon the assumption that clothing fashions are different from year to year. We achieve this research goal by formulating the fashion-year recognition problem into a clothing-fashion classification problem. There are three stages in this method. First, visual features are extracted from the human-body parts in an image. Second, the features are encoded into feature vectors, and fed into a learning machine to train a classification model. And third, the trained model is used to predict the fashion-year labels of new clothing images. It is worth noting that this study is different from existing studies related to clothing and fashions, which are aiming at clothing retrieval~\cite{hadi2015buy,liu2012street,liu2016deepfashion,liu2016fashion,wang2011clothes,yamaguchi2013paper}, clothing item parsing or segmentation~\cite{chen2012describing,chen2015deep,kiapour2014hipster,liu2014fashion,yamaguchi2012parsing,yang2014clothing} and clothing-style recognition~\cite{veit2015learning} and popularity recommendation~\cite{yamaguchi2014chic}.

In clothing fashion classification, human body segmentation is an indispensable step before extracting the clothing features. This is because, in most cases the clothing in the image occupies only a small part, which we call the foreground, while the image background is large and would bring significant influence to the classification. The classifier would be misled to predict the similar background rather than the similar foreground. Accurate human body masks will help in achieving the goal of fashion classification and prediction.
In this work, we present a deep learning based human body segmentation method based on a deep convolutional encoder-decoder architecture (SegNet)~\cite{segnet2015arxiv}. We improve the SegNet by fusing multi-scale convolutional features and build an end-to-end segment network while performing a peer-to-peer feature learning. Once human body masks are produced, we can conduct the fashion-year classification in the foreground region. In the classification stage, we design a lightweight deep convolutional neural network for fashion classification, and compare it with traditional classification methods that use some sophisticated hand-crafted features such as color, shape and texture descriptors.

The contributions of this work lie in three-fold. First, a dataset is constructed for fashion analysis and fashion classification. The dataset contains 9,339 fashion photographs with human-annotated foreground masks. Second, an end-to-end human body segmentation method is presented which fuses multi-scale features in deep architecture to produce accurate human body segmentation.
Third, a deep fashion classification network is proposed to predict the fashion year of a clothing.

The remainder of this paper is organized as follows.
Section~\ref{ssec:body} introduces the end-to-end human body segmentation method.
Section~\ref{ssec:fashion} describes in detail the clothing fashion classification network. Section \ref{sec:exp} reports
the experiment results, followed by a brief conclusion in
Section~\ref{sec:conc}.

\section{Body Segmentation}
\label{ssec:body}
Human body segmentation is a fundamental problem in computer vision, and it often plays as an important role in many applications such as gait recognition, human detection and re-identification. Traditionally, graph-cut-based methods have been widely used for human body segmentation~\cite{Boykov2001Interactive,Rother2004}. By calculating the probability of each pixel belongs to foreground and background, as well as evaluating the similarity between adjacent pixels, these methods build the graph cut model. An iterative computing strategy is often applied to reach an optimal segmentation. Generally, the graph-cut-based methods are effective when obvious difference exists between the foreground and background, but are ineffective when the foreground and background hold similar appearance. Moreover, the graph-cut-based methods often require human interactions to initialize.

The deep convolutional neural network (DCNN) has attracted wide attention as it shows astounding improved performance in solving many computer vision problems, e..g, image retrieval and classification~\cite{krizhevsky2012imagenet}. DCNN has also been used for edge and contour detection~\cite{wang2016edge,Shen2015CVPR} and image segmentation~\cite{rajchl2016deepcut}. Early segmentation networks follow the idea of image classification model. By cutting the image into small patches, these networks are trained to discern which category each patch belongs to, and then the classification results over the whole patches of an image can be combined and post-processed for image segmentation. These method do not provide end-to-end solutions for human body segmentation. Another kind of segmentation networks consist of only convolutional layers, where the traditional full connection layers are replaced with deconvolution layers. These networks are called fully convolutional networks. They often combine the deconvolution features with traditional low-level features to achieve the peer-to-peer learning. In fully convolutional networks, the global information can well be retained, and the network can be trained faster due to the largely decreased number of parameters. Typical networks of this kind for segmentation include the FCN~\cite{long2015fully}, U-net~\cite{Ronneberger2015U}, and V-net~\cite{Milletari2016V}, etc.

\begin{figure}[htb]
  \centering
\centerline{\includegraphics[width=1.0\linewidth]{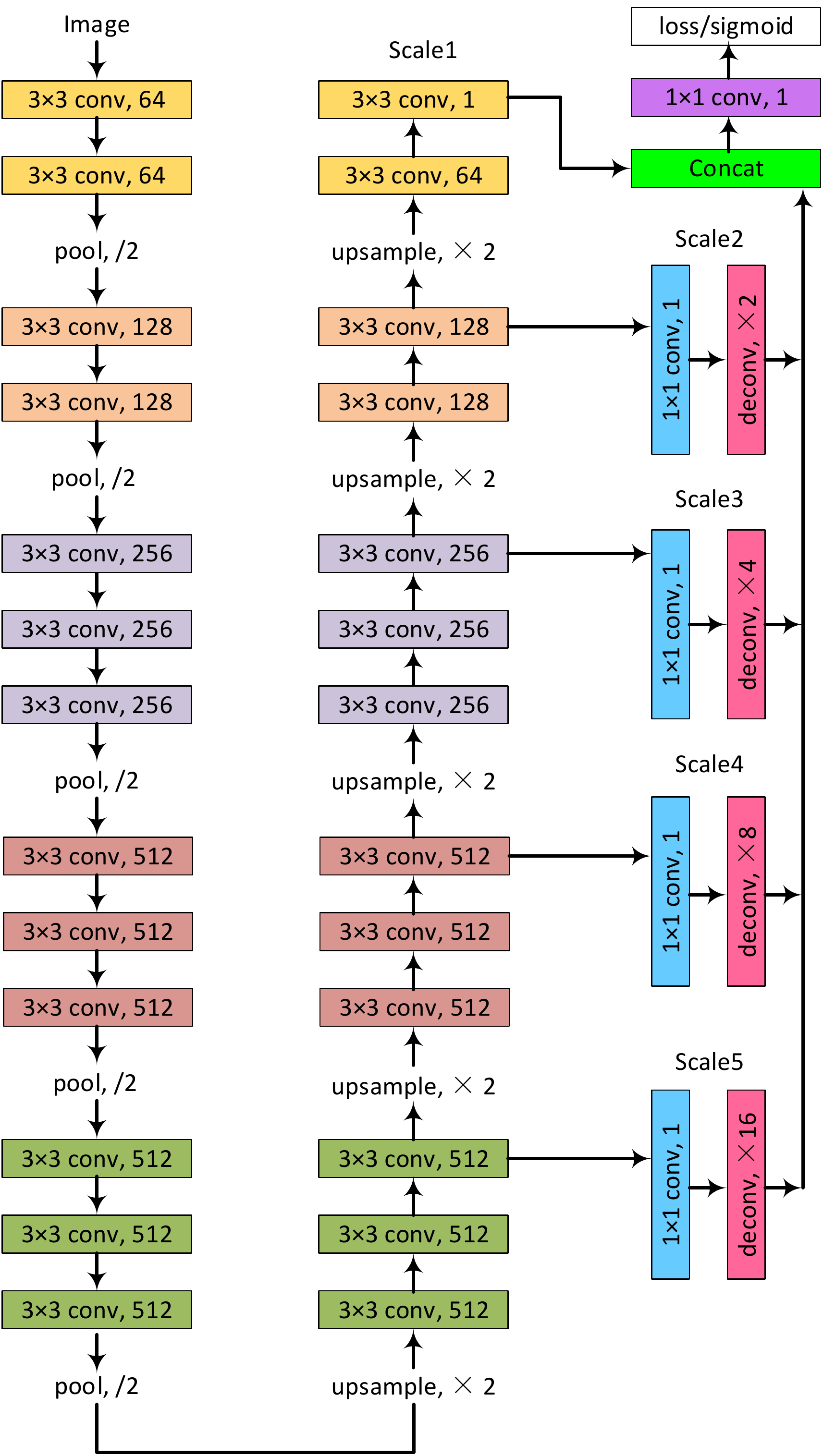}}
\caption{An overview of the proposed human-body segmentation network.}
\label{fig:arch}
\end{figure}

The proposed body segmentation network is adapted from a recently developed SegNet architecture~\cite{segnet2015arxiv}. SegNet is a deep convolutional architecture containing an encoder network and a corresponding decoder network, which is originally designed for semantic segmentation of drive-scene images. The encoder network is copied from the convolutional layers of the VGG16 network~\cite{Simonyan2014Very}, which consists of 13 convolutional layers and 5 pooling layers. The decoder network also has 13 convolutional layers, with each corresponding to a layer in the encoder network. The last convolution operation in decoder network produces a $c$-channel feature map, with $c$ the number of classes in the image segmentation task. After each convolution operation, a batch-normalization step is applied to the feature maps. In SegNet, there is one main stream that the feature maps in high level can only be obtained from the low level along this path. In this work, we add more connection branches to the high level to utilize multi-scale features in the decoder net. As shown in Fig.~\ref{fig:arch}, 1 $\times$ 1 convolution layers are added to the last convolution layers of the first four stages in the decoder net, followed by deconvolution layers to upsamling the feature maps to fit the original size of the input. Then, we concatenate all these deconvolution layers and the last convolution layer of main stream and connect another  1 $\times$ 1 convolution layer to fuse features of different scales. At last, we can obtain prediction maps that indicates the probability each pixels belong to foregrouth or background.

Suppose $S = \{(X^{n}, Y^{n}), n = 1,...,N \}$ be a training data set containing $N$ images, where $X^{n} = \{x_{i}^{(n)}, i = 1,...,I \}$ denotes the raw input image, $Y^{n} = \{y_{i}^{(n)}, i = 1,...,I, y_{i}^{(n)}\epsilon \{0,1\} \}$ denotes the ground-truth segmentation label map corresponding to $X^{n}$, $I$ denotes the number of pixel in the image, then the training work is to produce prediction maps approaching the ground truth.
Let's future suppose $W$ be the weight parameters in the proposed network, and for each image input, $F^{n} = \{f_{i}^{(n)}, i = 1,...,I \}$ be the output feature maps in the fused convolution layer, then the pixel-wise prediction loss can be computed as
\begin{equation}
\label{eq:one}
l(f_{i};W)=
\begin{cases}
  log(1-P(f_{i};W)), & \mbox{if} \ y_{i} = 0 \\
  log(P(f_{i};W)), & \mbox{otherwise},
\end{cases}
\end{equation}
where $f_{i}$ is the output feature map of the network in pixel $i$, and $P(f)$ is the standard sigmoid function, which transforms the feature map into a human-body probability map. Let $M$ be the batch size in the training, then the batch loss can be formulated as
\begin{equation}
\label{eq:two}
\mathcal{L}(W)=\sum_{n=1}^{M} \sum_{i=1}^{I} l(f_{i}^{(n)};W).
\end{equation}

\section{Clothing Fashion Classification}
\label{ssec:fashion}

After human-body segmentation, we can conduct clothing fashion classification based on removing the complex background. Classification is also a fundamental problem in computer vision. Traditionally, classification is achieved by feature extraction and classifier learning. Based on feature extraction, some efficient visual feature descriptors have been designed and applied widely, such as SIFT, LBP, and so on. For classifier learning, SVM (support vector machine) is often used to linear and nonlinear classification. Besides these handicraft feature descriptors, deep convolution neural networks, e.g., AlexNet~\cite{krizhevsky2012imagenet}, GoogLeNet~\cite{Szegedy2014Going}, VGG~\cite{Simonyan2014Very}, have achieved largely improved performance in general image classification based on large-scale real training images such as ImageNet~\cite{Deng2009ImageNet}.

While these deep methods have presented great advantage than the traditional methods, the deeply-learned features are supposed to hold a good capacity and high discriminative power in describing the clothing fashion. However, these classical classification network are designed for large scale object classification, which shows non-adaptive to this clothing fashion classification on comparatively smaller data set, what's more, with the deeper and larger network, the convergence is more difficult to be guaranteed.

\begin{table}[!htbp]
\scriptsize
\centering
\caption{Classification network architecture.\label{tbl:fashionet}}\vspace{1mm}
{
\begin{tabular}{c|c|c|c}
  \hline
  Layer & Kernel Size/Stride & Channels & Output Size  \\
  \hline
  \hline
  Convolution & 11 $\times$ 11 / 4  & 64 & 192 $\times$ 96 $\times$ 64   \\
   \hline
  Max Pooling & 3 $\times$ 3 / 2 & & 96 $\times$ 48 $\times$ 64 \\
   \hline
  Convolution & 5 $\times$ 5 / 1 & 128 & 96 $\times$ 48 $\times$ 128 \\
   \hline
  Max Pooling & 3 $\times$ 3 / 2  & & 48 $\times$ 24 $\times$ 128 \\
   \hline
  Convolution & 3 $\times$ 3 / 1 & 256 & 48 $\times$ 24 $\times$ 256 \\
   \hline
  Max Pooling & 3 $\times$ 3 / 2  & & 24 $\times$ 12 $\times$ 256 \\
   \hline
  Convolution & 3 $\times$ 3 / 1 & 384 & 24 $\times$ 12 $\times$ 384 \\
   \hline
   Max Pooling & 3 $\times$ 3 / 2  & & 12 $\times$ 6 $\times$ 384 \\
   \hline
  Convolution & 3 $\times$ 3 / 1 & 512 & 12 $\times$ 6 $\times$ 512 \\
   \hline
  InnerProduct & & & 256 \\
   \hline
  InnerProduct & & & 8 \\
  \hline
\end{tabular}}
\end{table}

In this work, we design a lightweight classification model. Our proposed fashion classification network includes five convolution layers and two fully connection layers. As can be seen in Table~\ref{tbl:fashionet}, the first convolution layer has a kernel of size 11 $\times$ 11, with a stride of 4 and the second convolution layer has a kernel of size 5 $\times$ 5, with a stride of 1. The kernel size of other three convolution layers is 3 $\times$ 3, and the stride of them is 1. Expect the last convolution layer, each convolution layer is followed by a max-pooling operator with kernel size of 3 $\times$ 3 and a stride of 2, which is convinced to avoid heavy computation cost and maintain the invariance effectively. Hence, after these pooling operators, the size of feature maps is reduced to 1/64 of the input size. We also add the Batch-Normalization layer and nonlinear activation layer - ReLu to these convolution layers. Batch-Normalization is convinced to help the network train stable and speed the train process. The unit numbers of the first fully connection layer is set to 256. As discussed in Section 1, we take the year in which the clothing is released by the fashion show as the class label, and train the networks to predict the fashion-year labels of new clothes. In this work, the unit numbers of last fully connection layer is set to 8 since our fashion classification dataset is collected from 8 years.
Noting that, although the proposed classification network have five convolution layers that is the same in AlexNet, it combines with more pooling operators of stride 2 and less fully-connection units, which leads to an obviously lighter architecture than AlexNet, GoogLeNet and VGG. The size of our model is about one seventeen of AlexNet. The classification performance will be represented in the experiment section.

\begin{table}[!htbp]
\scriptsize
\centering
\caption{FASHION8 contains 9,339 photographs of 14 world-famous brands in the recent 8 years.\label{tbl:fashion8}}\vspace{1mm}
{
\begin{tabular}{|c|c|c|c|c|c|c|c|c|c|c|}
\hline
 \multirow{1}{*}{Year} & {2008} & {2009} & {2010} & {2011} & {2012} & {2013} & {2014} & {2015} \\
  \hline \hline
  \multirow{1}{*}{Bott. V.} & 78 & 54 & 79 & 74 & 74 & 86 & 88 & 82 \\
  \hline
  \multirow{1}{*}{Burb. P.} & 80 & 87 & 78 & 83 & 79 & 85 & 85 & 85 \\
  \hline
  \multirow{1}{*}{Dior H.} & 51 & 79 & 80 & 75 & 79 & 79 & 82 & 80 \\
  \hline
  \multirow{1}{*}{Empo. A.} & 187 & 202 & 155 & 157 & 148 & 135 & 192 & 150 \\
  \hline
  \multirow{1}{*}{Erme. Z.} & 0 & 58 & 62 & 103 & 81 & 73 & 79 & 80 \\
  \hline
  \multirow{1}{*}{Givenchy} & 38 & 57 & 55 & 70 & 79 & 57 & 100 & 104 \\
  \hline
  \multirow{1}{*}{Gucci} & 87 & 89 & 89 & 85 & 73 & 79 & 75 & 72 \\
  \hline
  \multirow{1}{*}{Hermes} & 87 & 46 & 80 & 43 & 83 & 66 & 81 & 72 \\
  \hline
  \multirow{1}{*}{John V.} & 46 & 36 & 49 & 81 & 70 & 65 & 71 & 75 \\
  \hline
  \multirow{1}{*}{Louis V.} & 109 & 94 & 104 & 98 & 79 & 79 & 83 & 76 \\
  \hline
  \multirow{1}{*}{Neil B.} & 70 & 70 & 67 & 71 & 73 & 68 & 76 & 76 \\
  \hline
  \multirow{1}{*}{Prada} & 83 & 89 & 74 & 79 & 84 & 76 & 73 & 65 \\
  \hline
  \multirow{1}{*}{Thom B.} & 103 & 111 & 81 & 83 & 86 & 79 & 82 & 81 \\
  \hline
  \multirow{1}{*}{Versace} & 80 & 92 & 86 & 92 & 91 & 101 & 82 &  89 \\
  \hline
  \multirow{1}{*}{Sum} & {1099} & {1164} & {1139} & {1194} & {1179} & {1128} & {1249} & {1187} \\
  \hline
\end{tabular}}
\end{table}

\begin{figure*}[!ht]
\centering
\hspace{-0.2in}
\includegraphics[width=0.8\linewidth]{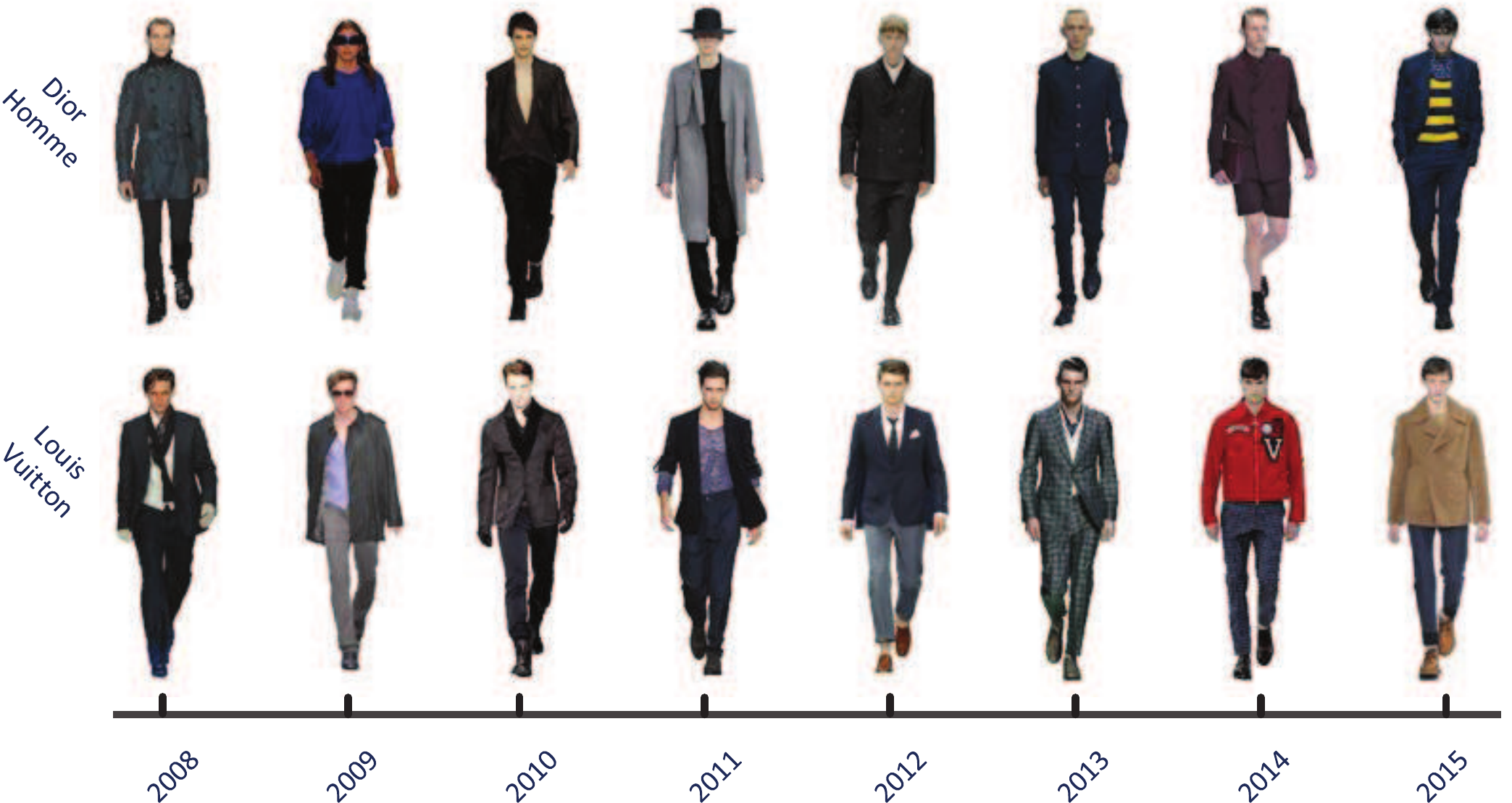}
\caption{Example photographs of two brands in FASHION8: Dior Homme and Louis Vuiton, from 2008 to 2015.}
\label{fig:fashion8}
\end{figure*}

\section{EXPERIMENTAL RESULTS} \label{sec:exp}

\subsection{Dataset} \label{sec:data}
We construct a dataset - FASHION8~{\footnote{It is available at https://sites.google.com/site/qinzoucn/documents}} - for training and testing our CNN models, as well evaluating the performance of the proposed methods. The FASHION8 dataset contains 9,339 photographs with resolution of 384$\times$768 pixels captured on the Fashion Shows from 2008 to 2015, where the details are shown in Table~\ref{tbl:fashion8}. Firstly, We use an interactive image segmentation tools to segment the human body from the fashion-show background. Images from the same year are taken as one category. Then, we randomly and proportionally choose 8,000 images from FASHION8 as the training set. The remaining 1,339 images are used as the test set.

\vspace{-2mm}
\subsection{Training} \label{sec:train}

\begin{figure}[t]
  \centering
  \centerline{\includegraphics[width=0.7\linewidth]{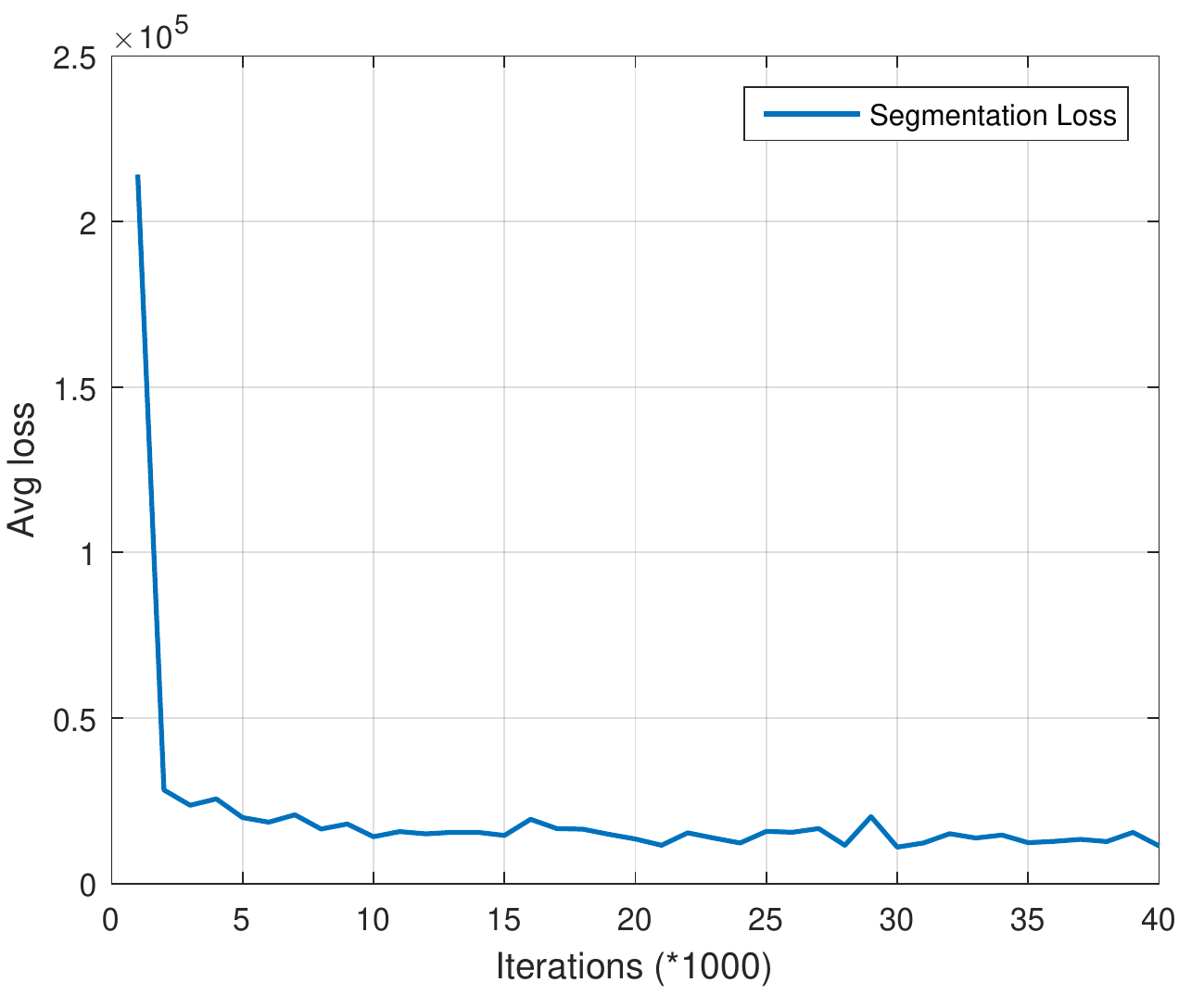}}
\caption{The loss curve of training human-body segmentation network. }
\label{fig:seg_loss}
\end{figure}

Segmentation: The weight parameters of $conv$ layers in the entire network are initialized by the `msra' method and the biases are initialized to 0. In training, the initial global learning rate is set to 1e-5 and will be divided by 10 after every 10k iterations. The momentum and weight decay are set to 0.9 and 0.0005, respectively. The stochastic gradient descent method (SGD) is employed to update the network parameters with mini-batch size of 2 in each iteration. We train the network with 40k iterations in total. The training loss curve is showed in Fig.~\ref{fig:seg_loss}.

Classification: Similar to the segmentation net, the weight parameters of $conv$ layers are initialized by the `msra' method and the biases are initialized to 0. We use a mini-batch size of 64, a total of 20,000 training iterations. The base learning rate is initialized as $1e^{-3}$, the momentum and weight decay are set to 0.9 and 0.001. The training loss curve is showed in Fig.~\ref{fig:seg_loss}.

We implement our network using the publicly available Caffe~\cite{Jia2014Caffe} which is well-known in this community and all experiments in this paper are performed by using an NVIDIA TITAN X GPU.

\begin{figure}[t]
  \centering
  \centerline{\includegraphics[width=0.7\linewidth]{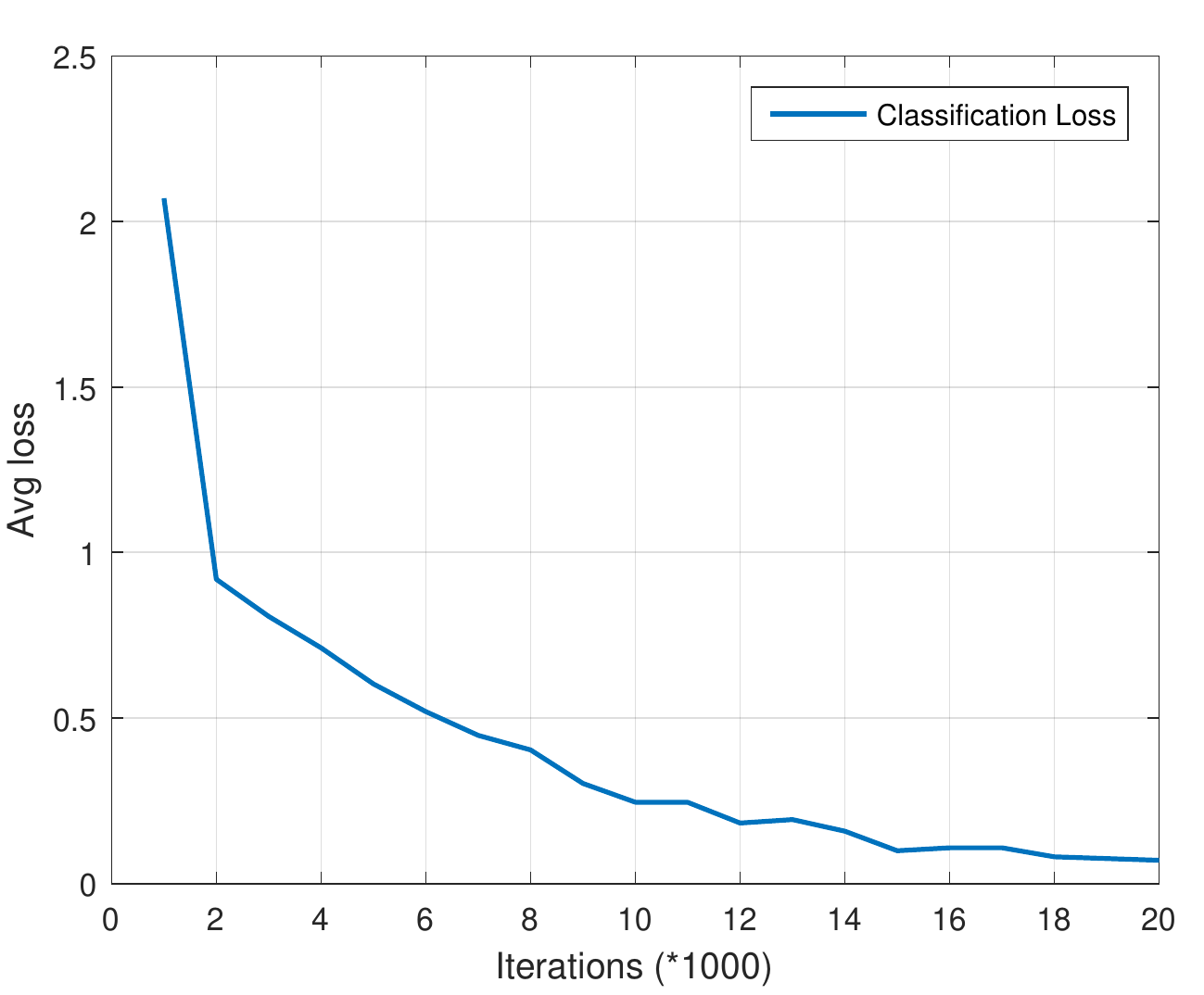}}
\caption{ The loss curve of training fashion clothing classification network. }
\label{fig:class_loss}
\end{figure}

\begin{figure}[t]
  \centering
  \centerline{\includegraphics[width=0.9\linewidth]{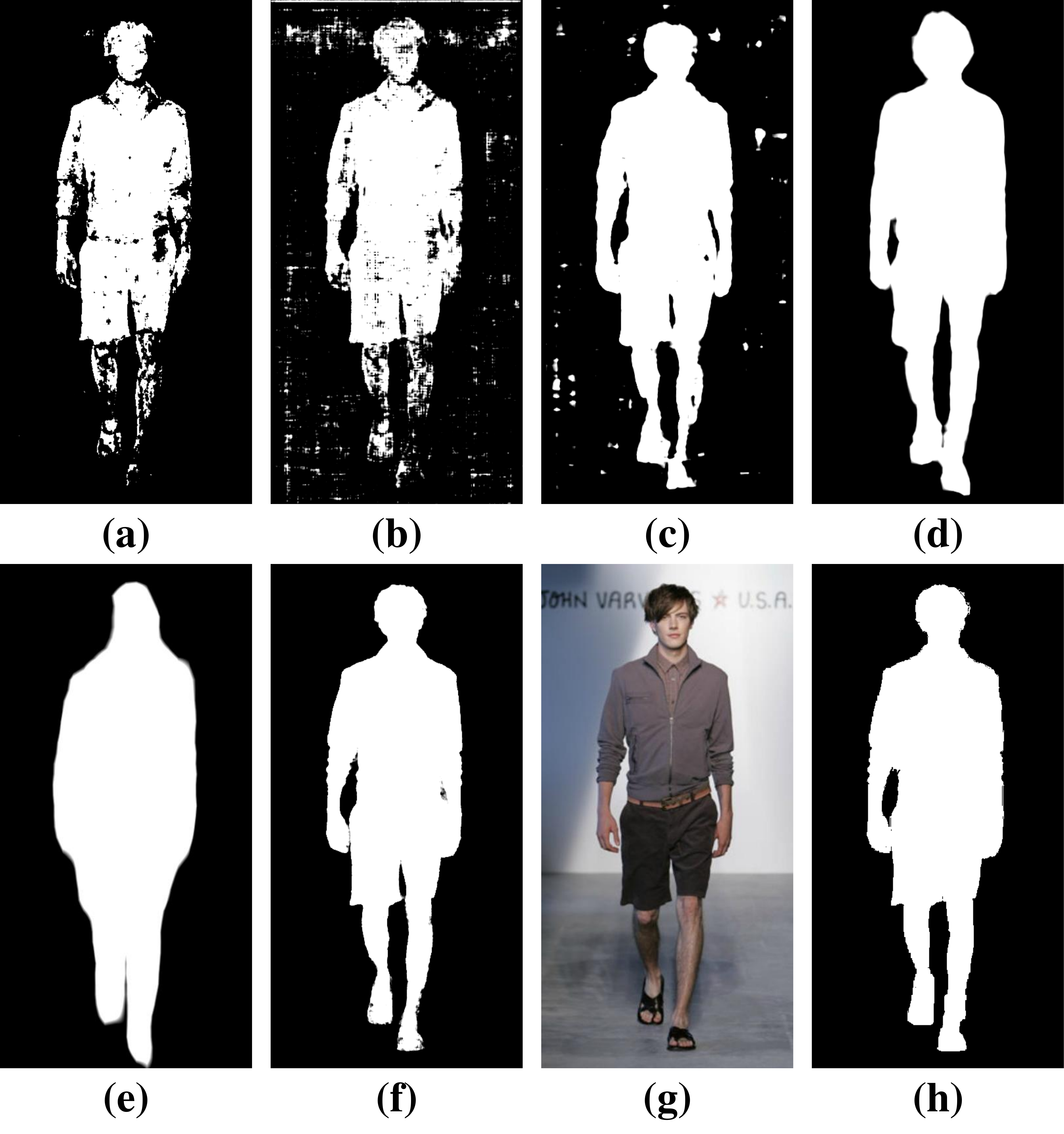}}
\caption{Prediction results of single scale features and the fuse features. (a)-(e) are the output prediction from scale 1 to 5, respectively. (f) is the fuse prediction results, (g) is the input image, and (h) is the ground-truth human body mask.}
\label{fig:ms_results}
\end{figure}

\begin{figure*}[!htbp]
\centering
\begin{minipage}[]{0.96\linewidth}
  \centering
  \centerline{\includegraphics[width=1\textwidth]{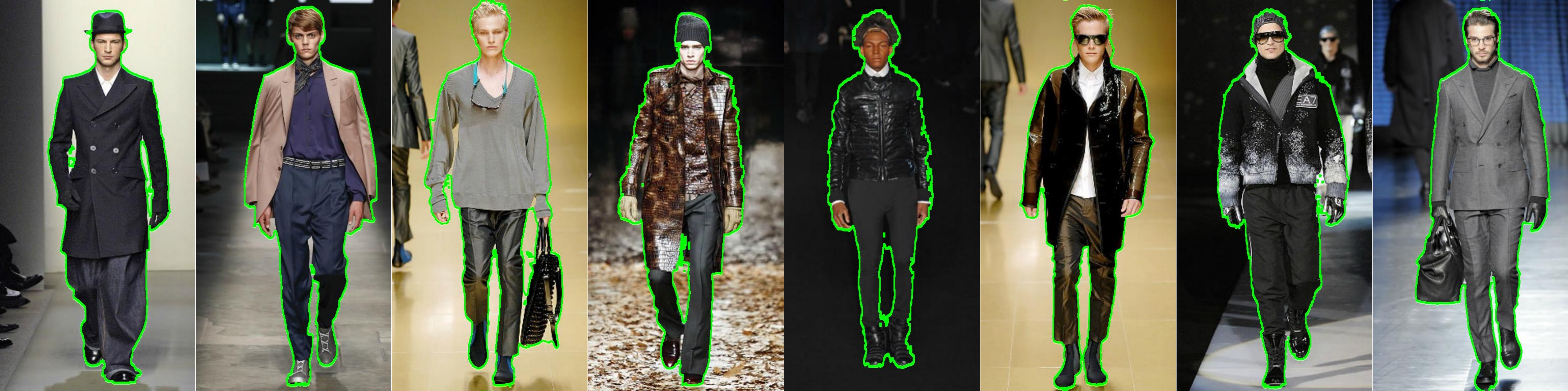}}
\end{minipage}
\caption{Body segmentation results produced by the proposed method. Green lines show the boundaries of the segmented human bodies.}
\label{fig:bodyseg}
\end{figure*}

\subsection{Results} \label{sec:result}

\begin{figure}[t]
  \centerline{\includegraphics[width=0.98\linewidth]{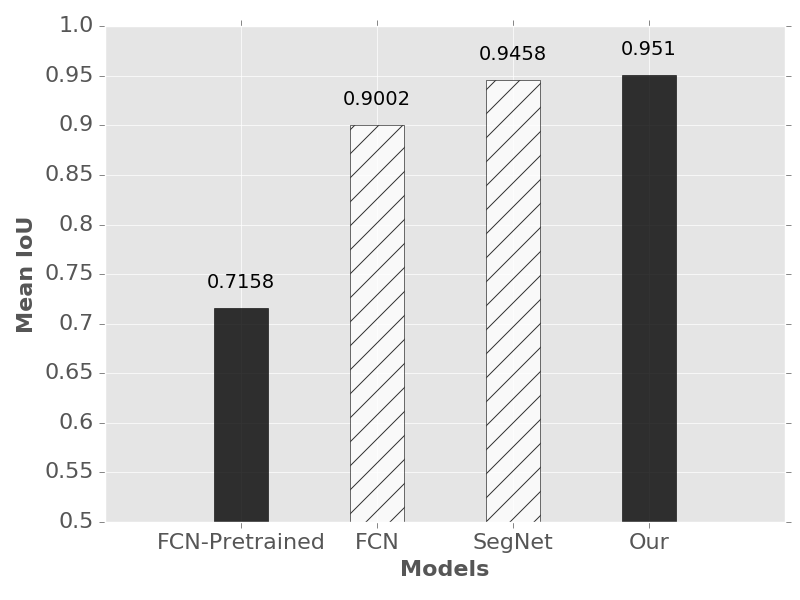}}
\caption{Comparison of the mean $IoU$s obtained by the different models on FASHION8 testing set.}
\label{fig:mean-iou}
\end{figure}

{\textbf{Body Segmentation.}} The metric $IoU$ is employed to quantify the accuracy of the segmentation for one single image,
\begin{equation}\label{eq:iou}
    {IoU}=\frac{\mathcal{A}\bigcap \mathcal{B}}{\mathcal{A}\bigcup \mathcal{B}},
\end{equation}
where $\mathcal{A}$ denotes the ground-truth segmentation, and $\mathcal{B}$ denotes the segmentation result. Then the accuracy on the a set of $N$ images is defined as,
\begin{equation}\label{eq:meaniou}
    mean IoU=\frac{1}{N} \sum_{k=1}^N IoU_k.
\end{equation}

Figure~\ref{fig:ms_results} displays the segmentation results of single scales and the fused layer produced by the proposed method. We can see that the smaller scale produces more detailed result (e.g., Fig.~\ref{fig:ms_results}(a)) and the larger scale produce more blur results (e.g., Fig.~\ref{fig:ms_results}(e)). While combining the characteristics of all these scales, the fused layer can produce accurate body segmentation. Note that, the smaller scale in this work does not mean the low resolution, instead, it means high resolution.

Figure~\ref{fig:bodyseg} displays some segmentation results produced by the proposed method. Figure~\ref{fig:mean-iou} shows the mean $IoU$ of body segmentation results on the test set of FASHION8. While the published FCN network is trained on VOC'11 and the VOC'11 dataset contains an objective category of human, we can directly use the pre-trained FCN model for human body segmentation, the mean $IoU$ of which is 0.7158, much lower than that of the proposed method. Based on this model, we fine-tune the FCN network on our training set, and obtain a mean $IoU$ of 0.9002. The original SegNet is trained for road scene segmentation. For our body segmentation, we retrain the SegNet model and obtain a mean $IoU$ of 0.9458. We train the proposed segmentation network from scratch and get the best segmentation results among these compared methods, with a mean $IoU$ of 0.9510. This high accuracy body segmentation is important for the following fashion classification.

\vspace{1mm}
\noindent
{\textbf{Clothing Fashion Classification.}}
In this part, we evaluate the proposed classification network on the foreground region, i.e., human body, of the FASHION8 dataset. For comparison, several sophisticated feature descriptors are employed in our study, e.g., the `SIFT-IFV' which encodes shape features using IFV~\cite{perronnin2010improving} over dense SIFT descriptors~\cite{lowe2004distinctive}, the `COLOR-RCC' which encodes color features using a regional color co-occurrence matrix~\cite{zou2015discriminative}, and the `PRICoLBP' which encodes texture features using LBPs~\cite{Qi2014Pairwise}. Besides, we also compare our model with the classical AlexNet.

\begin{table}[htb]
\caption{Classification results.}
\label{tbl:classification} \vspace{1mm}
\small
\centering
\begin{tabular}{|c|c|c|}
\hline
{{Methods}} & Accuracy & Feature Dimensions\\
\hline\hline
COLOR-RCC(256) & 0.716 & 32896\\
\hline
PRICoLBP & 0.516  & 17700\\
\hline
SIFT-IFV (512)&  0.774 & 71749 \\
\hline
AlexNet &  0.698  & --\\
\hline
Our &  0.805 & -- \\
\hline
\end{tabular}
\end{table}

\begin{figure}[t]
  \centering
  \centerline{\includegraphics[width=0.98\linewidth]{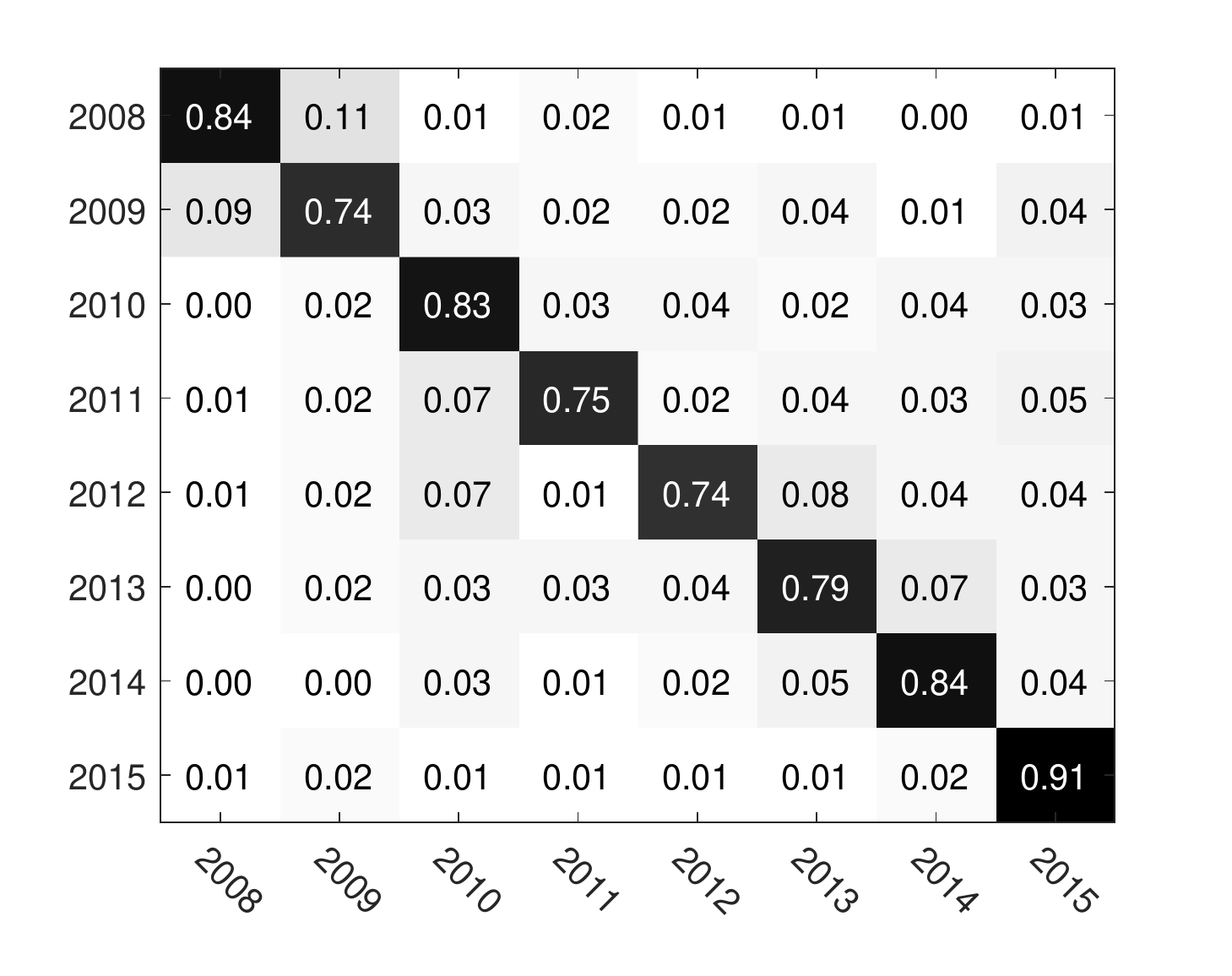}}
\caption{ The confusion matrix of classification results obtained by our method.}
\label{fig:co-matrix}
\end{figure}

Table 2 shows the classification results on the testing set. Among the traditional feature descriptors, SIFT-IFV(512), which uses 512 Gaussian models in IFV, achieves the highest classification accuracy of 0.774. The accuracy of COLOR-RCC(256), which uses 256 color codes, is 0.716, and the accuracy of PRIColBP is 0.516.
We train the AlexNet on the FASHION8 dataset and get an accuracy of 0.698, which is slightly lower than that of SIFT-IFV(512). This low accuracy may be because the AlexNet trained on a small-scale dataset would easily to be overfitting. The proposed network obtains an accuracy of 0.805, which demonstrates that the proposed model can effectively capture the information of the clothing fashion and distinguish clothing fashion between different years. Figure~\ref{fig:co-matrix} shows the confusion matrix on the results obtained by the proposed methods. It can be observed that, two years with shorter interval generally hold more confusable clothing fashions.

\section{CONCLUSION}
\label{sec:conc}
In this paper, the fashion-time recognition problem was mapped into a clothing-fashion classification problem. Specifically, an end-to-end human body segmentation method was proposed to produce accuracy foreground mask for clothing-fashion classification. The proposed segmentation net was an improved convolutional encoder-decoder architecture and achieved higher mean $IoU$ rate for human body segmentation while trained on the FASHION8 dataset. In fashion classification, a lightweight network was proposed for predicting which year the clothing belongs to, and outperformed traditional handicraft features based methods and the AlexNet on this clothing-fashion classification task. The results demonstrated the effectiveness of the proposed body segmentation method and the fashion classification method.

\section*{Acknowledgement}
This research was supported by the National Natural Science Foundation of China under grant No.~61301277 and No.~61572370, Key Research Base for Humanities and Social Sciences of Ministry of Education Major Project under grant No.~16JJD870002, and the Open Research Fund of State Key Laboratory of Information Engineering in Surveying, Mapping and Remote Sensing under Grant 16S01.

\bibliographystyle{IEEEtran}
\bibliography{refs}

\end{document}